# Large-Scale Domain Adaptation via Teacher-Student Learning

*Jinyu Li, Michael L. Seltzer, Xi Wang, Rui Zhao, and Yifan Gong*

Microsoft AI and Research, One Microsoft Way, Redmond, WA 98052

{jinyli; mseltzer; xwang; ruzhao; ygong}@microsoft.com

## Abstract

High accuracy speech recognition requires a large amount of transcribed data for supervised training. In the absence of such data, domain adaptation of a well-trained acoustic model can be performed, but even here, high accuracy usually requires significant labeled data from the target domain. In this work, we propose an approach to domain adaptation that does not require transcriptions but instead uses a corpus of unlabeled parallel data, consisting of pairs of samples from the source domain of the well-trained model and the desired target domain. To perform adaptation, we employ teacher/student (T/S) learning, in which the posterior probabilities generated by the source-domain model can be used in lieu of labels to train the target-domain model. We evaluate the proposed approach in two scenarios, adapting a clean acoustic model to noisy speech and adapting an adults' speech acoustic model to children's speech. Significant improvements in accuracy are obtained, with reductions in word error rate of up to 44% over the original source model without the need for transcribed data in the target domain. Moreover, we show that increasing the amount of unlabeled data results in additional model robustness, which is particularly beneficial when using simulated training data in the target-domain.

## 1. Introduction

The success of deep neural networks [1][2][3][4][5] relies on the availability of a large amount of transcribed data to train millions of model parameters. However, deep models still suffer reduced performance when exposed to test data from a new domain. Because it is typically very time-consuming or expensive to transcribe large amounts of data for a new domain, domain-adaptation approaches have been proposed to bootstrap the training of a new system from an existing well-trained model [6][7][8][9]. These supervised methods still require transcribed data from the new domain and thus their effectiveness is limited by the amount of transcribed data available in the new domain. Although unsupervised adaptation methods can be used by generating labels from a decoder, the performance gap between supervised and unsupervised adaptation is large [7].

In this work, we propose an approach to domain adaptation that does not require transcriptions but instead uses a corpus of unlabeled parallel data, consisting of pairs of samples from the source domain of the well-trained source model and the target domain. There are many important scenarios in which collecting a virtually unlimited amount of parallel data is relatively simple. For example, to collect noisy or reverberant data from a particular set of environments, speech can be captured simultaneously using a close-talking microphone and a microphone located at a distance from the user. Such a collection effort can also be simulated by acoustically replaying a pre-existing corpus of high signal-to-noise ratio speech files in the target environment or by digitally simulating the target environment offline [10][11].

To perform adaptation without the use of transcriptions, we propose to use teacher/student (T/S) learning. In T/S learning, the data from the source domain are processed by the source-domain model (teacher) to generate the corresponding posterior probabilities or soft labels. These posterior probabilities are used in lieu of the usual hard labels derived from the transcriptions to train the target (student) model with the parallel data from the target domain. With this approach, the network can be trained on a potentially enormous amount of training data and the challenge of adapting a large-scale system shifts from transcribing thousands of hours of audio to the potentially much simpler and lower-cost task of designing a scheme to generate the appropriate parallel data.

The proposed approach is closely related to other approaches for adaptation or retraining that employ knowledge distillation [12]. In these approaches, the soft labels generated by a teacher model are used as a regularization term to train a student model with conventional hard labels. Knowledge distillation was used to train a system on the Aurora 2 digit recognition task [13], using the clean and noisy training sets [14]. In [15] it was shown that for the multi-channel CHiME-4 task [16], soft labels could be derived using enhanced features generated by a beamformer then processed through a network trained with conventional multi-style training [17]. However, it is unclear whether this approach is superior to simply using the enhanced features for the recognition at test time as well. Knowledge distillation was also used to adapt an acoustic model to new dialects using a small adaptation corpus [18]. In all cases, the soft labels provided by the teacher network regularized the conventional training of the student network using hard labels derived from transcriptions. Thus, the use of additional unlabeled training data was not possible.

In contrast, the proposed approach forgoes the need for hard labels from the data in the new domain entirely and relies solely on the soft labels provided by the parallel corpus and well-trained source model. This allows the use of a significantly larger set of adaptation data which adds robustness to the resulting model. In this work, for example, the unlabeled training data represents an *order of magnitude* more acoustic data than was used to create the well-trained source model. We evaluate the proposed approach in two scenarios, adapting a clean acoustic model to noisy speech and adapting an adults' speech acoustic model to children's speech. We show that the resulting noisy speech model can obtain performance significantly better than multi-condition training with far better robustness to unseen noise conditions. Significant reduction in word error rate (WER) is obtained on children's speech when no children's speech is present in the original source model.

## 2. T/S learning for domain adaptation

In this section, we present T/S learning as a general framework for domain adaptation using unlabeled data. We propose to directly minimize the Kullback–Leibler (KL) divergence between the output distribution of the student network and the teacher network by leveraging large amounts of unlabeled parallel data as shown in Figure 1. We denote the posterior distribution of the teacher and student networks as $P_T(s|x_{src})$ and $P_S(s|x_{tgt})$, respectively. $x_{src}$ and $x_{tgt}$ are the source and target inputs to the teacher and student networks, respectively. The KL divergence between these two distributions is

$$\sum_f \sum_i P_T(s_i|x_{src,f}) log\left(\frac{P_T(s_i|x_{src,f})}{P_S(s_i|x_{tgt,f})}\right), \quad (1)$$

where $s$ indicates senone, $i$ is the senone index and $f$ is the frame index. This formulation takes both the source data $x_{src}$ and the target data $x_{tgt}$, differing from the original T/S formulation in [19] which takes the same data for teacher and student networks.

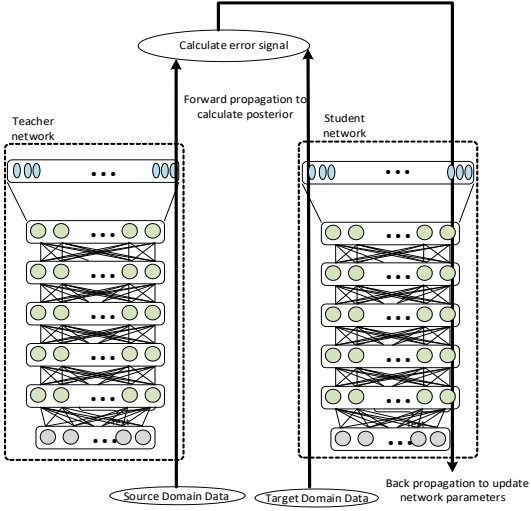

Figure 1: *The flow chart of teacher-student learning using parallel data for domain adaptation*

To learn a student network that approximates the given teacher network, only the parameters of the student network needs to be optimized. Minimizing the above KL divergence is equivalent to minimizing

$$-\sum_f \sum_i P_T(s_i|x_{src,f}) log P_S(s_i|x_{tgt,f}) \quad (2)$$

because $P_T(s_i|x_{src,f}) log P_T(s_i|x_{src,f})$ has no impact to the student network parameter optimization. The training steps of a student network guided by a teacher network which is well-trained with source-domain transcribed data are:

1. Clone the student network from the teacher network.
2. Use *parallel unlabeled* source data $x_{src}$ and target data $x_{tgt}$ to train the student network with the following steps.
   a. For each mini-batch, do forward propagation of teacher network using $x_{src}$ and student network using $x_{tgt}$ to calculate $P_T(s_i|x_{src,f})$ and $P_S(s_i|x_{tgt,f})$.
   b. Calculate the error signal of Eq. (2), and then do back propagation for the student network.
   c. Repeat Step 2.a & 2.b until convergence.

The advantage of the proposed method is that training of the student network doesn't need any transcription as long as we have parallel data because its supervision signal $P_T(s_i|x_{src,f})$ is obtained by passing the source data through the teacher network. Thus, the student network trained optimizing Eq. (2) can exploit unlimited parallel training data. Increased training data provides better coverage of the acoustic space, such that the student network in the target domain behaves very similarly to the well-trained teacher network in the source domain.

## 3. Parallel corpus generation

T/S learning for domain adaptation relies on the availability of a parallel corpus of source and target data, which consists of unlabeled real or the simulated training pairs from the source and target domains, respectively. In this study, we explore two domain-adaptation scenarios: 1) adapting from clean to noisy environments; 2) adapting from adults to children speech.

It is quite straightforward to collect a parallel corpus of clean and noisy speech. Real paired examples can be obtained by replaying clean speech in a noisy environment. Simulated examples can be obtained by digitally mixing the clean speech with noise. Then, Eq. (2) can be re-written as

$$-\sum_f \sum_i P_T(s_i|x_{clean,f}) log P_S(s_i|x_{noisy,f}).$$

Obtaining a parallel corpus of adult and child speech is more challenging. It is very hard to synchronize natural speech from the adults and children so that the resulting samples are synchronized. We opt for a voice transformation approach to simulate children's speech, using formant-based frequency warping. We adopt bilinear frequency warping on the adults speech spectrum and reconstruct the signal with higher pitch. The bilinear transform [20][21] produces the frequency transformation as $\omega_{new} = \omega + 2 \arctan\left[\frac{-\alpha \sin(\omega)}{1+\alpha \cos(\omega)}\right]$, where $\omega$ denotes the frequency and the parameter $\alpha$ decides the warping factor. For general voice conversion, there is typically a mapping from source to target speech which can be used to calculate the $\alpha$ value from vowel segments. In this work, we simply select a warping factor of 0.1 which moves the format frequencies higher. Thus, we are not performing voice conversion *per se* since there is not a specific voice target. In this scenario, Eq. (2) can be re-written as

$$-\sum_f \sum_i P_T(s_i|x_{adults,f}) log P_S(s_i|x_{children,f}).$$

## 4. Experimental evaluation

The proposed methods are evaluated using several tasks. The baseline acoustic model for all experiments is a 4-layer LSTM-RNN [22] with 5976 senones trained with the cross-entropy criterion. LSTM-RNNs have been shown to be superior than the feed-forward DNNs [22][23]. Each LSTM layer has 1024 hidden units and the output size of each LSTM layer is reduced to 512 using a linear projection layer. There is no frame stacking, and the output HMM state label is delayed by 5 frames as in [22]. The input feature is the 80-dimension log-filter-bank feature. The transcribed data used to train the baseline acoustic model comes from 375 hours of US-English Cortana audio.

### 4.1 Noisy Cortana task

In this series of experiments, we investigate adaptation of a clean acoustic model to noisy speech [24][25]. We consider the original Cortana data in the source domain as the clean source data. While this data is not noise-free, Table 2 shows that about 80% of the data is at SNRs higher than 20 dB. The target domain is a simulated noisy Cortana task which is obtained by digitally adding the ETSI noise [26] to the original Cortana data with SNRs from 5 to 20 dB. The noise-adding process assumes the Cortana data is clean, i.e. noise-free. The test set is extracted from Cortana live traffic from mobile phones, containing around 72,000 words, which guarantees the statistical significance of reported improvement. A trigram language model is used for decoding with around 8 million n-grams.

Table 1: *WERs (%) on Cortana test sets. The first column indicates the training set for the teacher model. The second column describes the unlabeled training set for the student model. If none, the teacher model is used for evaluation. The third column shows the environment, either original or noisy.*

| Train Teacher (transcribed) | Train Student (unlabeled pairs) | Cortana evaluation condition | |
|---|---|---|---|
| | | original | noisy |
| original 375h | none | 15.62 | 18.80 |
| noisy 375h | none | 16.58 | 17.34 |
| original 375h | original-noisy 375h | 15.32 | 16.66 |
| original 375h | original-noisy 3400h | **15.17** | **16.11** |

Table 2: *WER (%) breakdown for original Cortana condition*

| SNR | <5 db | [5, 20]db | [20,35]db | >35db |
|---|---|---|---|---|
| Word Count | 155 | 15568 | 31652 | 25015 |
| Baseline | 23.87 | 15.71 | 15.46 | 14.7 |
| T/S 3400h | 20.65 | 15.31 | 14.9 | 14.44 |

Table 1 shows the WER for different systems. The original evaluation is with the original Cortana test data, and the noisy evaluation is with the noise-added Cortana data. The baseline LSTM obtained 15.62% WER in the original source condition and increases to 18.80% in the noisy target condition. We then add noise to the baseline 375h of training data to train a standard multi-condition LSTM model. This model improves the WER in noisy condition to 17.34% WER but degrades the WER in original condition to 16.58%. We next evaluate the proposed T/S framework using the parallel corpus of 375 hours without using the transcription, with the original and noisy data as the inputs to the teacher and student networks, respectively. In this case, the student network obtains a WER of 16.66%, significantly outperforming the standard multi-condition model. If we increase the size of the parallel training data to 3400 hours the WER further improves to 16.11%, showing the advantage of exploiting a large amount of unlabeled data. The 16.11% WER obtained in the target domain is very close to the 15.62% WER obtained by the source (teacher) model on the original Cortana task. This indicates that the T/S learning is effective in that the behavior of the student network in the target domain is approaching that of the teacher network in the source domain. Note that we did not see any improvement using knowledge distillation with the hard labels derived from transcriptions. This is consistent with the findings reported in [15][27].

At first glance, it is surprising that the student models trained from 375h and 3400h parallel data outperform the teacher model on the original source condition. To understand this behavior, we broke down the WERs of the original test set into different SNR levels by running an automatic SNR detector on every utterance. The resulting WERs are shown in Table 2, for the baseline source model and the adapted T/S model. Note that few utterances failed to generate SNR results, and hence the weighted average WER in Table 2 is slightly different from the WER in Table 1. As the table indicates, some of the original Cortana utterances are already noisy. The student model clearly wins for SNR levels below 35dB most likely because the simulated noisy utterances for the parallel training have [5, 20]dB SNRs and the soft-target learning may extrapolate well beyond that SNR range. It is still worth investigating why the student model even wins for SNRs larger than 35dB although the gap is small.

### 4.2 CHiME-3 task

We next investigated how the models learned in Section 3.1 behave in a highly-mismatched test environment. The mismatched task we choose is CHiME-3 [28], which contains the Wall Street Journal (WSJ) utterances recorded in real noisy environments. The single-channel far-field noisy speech (the 5th microphone channel) is used for evaluation. WSJ 5K word 3-gram language model is used for decoding. The CHiME-3 test set and the Cortana parallel training data are mismatched in terms of task, speaking style, microphone, environment etc. Also, the noisy speech in the Cortana parallel data is simulated while the CHiME-3 test set is real speech.

Table 3 compares WERs from different models. The baseline model trained on 375h of Cortana data obtains a WER of 23.16%. Because of the significant mismatch between Cortana and CHiME-3, both the model trained with 375h of noisy transcribed data and the student model trained with parallel 375h original-noisy data fail to improve performance. However, the student model trained with parallel 3400h of parallel original/noisy Cortana data improves the WER to 19.89%, a 14% relative WER reduction. This gain results from significantly increasing the amount of parallel training data which helps the student model cover much more of the acoustic space.

Although the essence of T/S learning is using very large amount of unlabeled data so that the student's behavior in the target domain can approach the teacher's behavior in the source domain, we also want to evaluate the performance of T/S learning when only limited parallel data is available. To that end, we used the parallel data from CHiME-3 training set to adapt the baseline 375h Cortana model. Now, the source data comes from the clean CHiME-3 data, while the noisy target data comes from different sources by combining the real and simulated data from one or more microphones as shown in Table 4. The numbers of real and simulated utterances in each channel are around 2k and 7k, respectively. The T/S learning using either the 2k parallel clean-real channel 5 or the 7k clean-simulated channel 5 utterances can reduce the WER from 23.16% to just under 16%. The results show that using the real data is most effective, but if the real data is unavailable more simulated data can be used. By combining both the real and simulated channel 5 data as the input to the student network, T/S learning can further reduce the WER to 13.77%. Then, with more data from the other microphones, T/S learning can get further improvement. The final student model which was trained with the real and simulated data from all channels get the 12.99% WER, about 44% relative WER reduction over the original source model. This improvement is much larger than what can be obtained from the traditional feature mapping [29][30] and mask learning [31][32] methods. In [33], we proposed advanced models to improve the feature mapping and mask learning methods, but can only obtain 25% relative WER reduction, far below the improvement obtained from the T/S learning in this work.

Table 3: *WERs (%) on Chime 3 test sets using Cortana data. The columns have the same meaning as in Table 1.*

| Train Teacher (transcribed) | Train Student (unlabeled pair) | WER |
|---|---|---|
| original 375h | None | 23.16 |
| noisy 375h | None | 24.51 |
| original 375h | original- noisy 375h | 23.67 |
| original 375h | original- noisy 3400h | **19.89** |

Table 4: *WERs (%) on Chime 3 test sets using Chime 3 data. The source data is the clean data. The target data comes from different noisy sources.*

| The noisy target data in the pair comes from | | | | |
|---|---|---|---|---|
| Real channel 5 | Simulated channel 5 | Other real channels | Simulated other channels | WER |
| Y | N | N | N | 15.88 |
| N | Y | N | N | 15.73 |
| Y | Y | N | N | 13.77 |
| Y | Y | Y | Y | **12.99** |

### 4.3 Children's speech

In this section, we explore the T/S learning method for children's speech recognition which is important to home entertainment applications [34][35]. We first run a DNN gender classifier to determine the percentage of the male adults, female adults, and children in the 375h transcribed data as: 70.5%, 25.3%, and 4.2%, respectively. We remove both children and female adults' data from the training set, as some female adults' data and children are acoustically similar. We then train a baseline LSTM model from only the adult male data, with the same structure as the baseline LSTM model in Section 3.1. Table 5 gives the model evaluation results on children's utterances, recorded from boys and girls. This adult male LSTM model preforms very poorly, with 39.05% and 34.16% WER, on the girls and boys test sets, respectively.

Then, we use the bilinear transform described in Section 2.2 to transform adult speech into simulated children's speech. The quality of the transformation seems to be an issue as the DNN gender classifier only labels 6.8% of the transformed utterances as children's speech. Thus, we only use those 6.8% of the transformed utterances as the target data, with the corresponding adult utterances as the source data. After T/S learning, the student model significantly improves the WER of girls' speech to 25.03%, and moderately improves the WER of boys' speech to 32.32%. The student model gets further improvement by extending the training set to the 3400h training set and selecting the portion of data that DNN gender classifier labels as children's speech. With this additional parallel data, the student model achieves a WER of 21.19% and 31.89% on the girls' and boys' test sets, respectively.

Finally, we evaluate whether the LSTM trained with all data (adults and children) can still benefit from T/S learning. Table 6 shows this baseline LSTM model has 18.38% and 22.98% WER on the girls and boys test sets, respectively. Therefore, this LSTM model is much better in handling children's speech. Using the gender DNN trusted transformed utterances from 375h data, the student model cannot improve anymore. But using the gender DNN trusted transformed utterances from 3400h data can improve the girls' speech to 16.65% WER, but still significantly degrades the WER of boys' speech.

Table 5: *WERs (%) of models initiated from Male LSTM on children's speech tasks.*

| Model | girls | boys |
|---|---|---|
| Adult male data from 375h transcribed | 39.05 | 34.16 |
| Target data: transformed children from 375h unlabeled | 25.03 | 32.32 |
| Target data: transformed children from 3400h unlabeled | **21.19** | **31.89** |

Table 6: *WERs (%) of models initiated from LSTM trained with male, female, children data on children's speech tasks.*

| Model | girls | boys |
|---|---|---|
| All data (male, female, children) from 375h transcribed | 18.38 | **22.98** |
| Target data: transformed children from 375h unlabeled | 18.86 | 30.24 |
| Target data: transformed children from 3400h unlabeled | **16.65** | 29.20 |

Both Tables 5 & 6 show the advantage of using a large amount of data. More parallel data means that the student model can explore more of the acoustic space, resulting in good adaptation performance. In Table 5, the target domain data (children's speech) is not observed in the source domain (adult male speech). Therefore, it is very easy to observe a gain. However, in Table 6, the target domain data is already well modeled by the source model, and hence it is more challenging to get improvement with simple voice conversion approaches. We have listened to the transformed utterances and found that it is relatively easy to obtain girls' speech via voice transformation but harder to create accurate examples of boys' speech. Therefore, the student models perform poorly when evaluated on boys' speech. We expect further improvements with a better voice transformation process.

## 5. Conclusions

In this study, we explore the large-scale domain adaptation using the T/S learning framework. To learn a deep network in a target domain without labeled data, we minimize the KL divergence of the output distribution between the source domain model with source data and the target domain model with target data. Different from the distillation framework which needs transcribed data, the T/S learning method relies on parallel unlabeled data which is easier to obtain. By increasing the size of the unlabeled parallel training data, the behavior of student network in the target domain is very close to that of teacher network in the source domain.

Evaluated with the noisy Cortana task, the T/S learning student model can achieve a 16.11% WER, very close to the 15.62% WER obtained by the source model on the original Cortana task. On the CHiME-3 task, the student model gets up to 44% relative WER reduction over the source model. On the children's speech recognition task, the student model improves the WER of girls' speech significantly, but it is very challenging to improve the WER of boys' speech due to the limitations of the voice conversion method we employed. All experiments demonstrated that increasing the amount of unlabeled data results in additional model robustness, which is particularly beneficial when using simulated data in the target-domain.